# Inpainting borehole images using Generative Adversarial Networks

Rachid Belmeskine
Convergaince Corporation
One Broadway, Kendall Square
Cambridge, MA 02142
rachid@convergaince.com

Abed Benaichouche
Convergaince Corporation
One Broadway, Kendall Square
Cambridge, MA 02142
abed@convergaince.com

*Abstract*— In this paper, we propose a GAN-based approach for gap filling in borehole images created by wireline microresistivity imaging tools. The proposed method utilizes a generator, global discriminator, and local discriminator to inpaint the missing regions of the image. The generator is based on an auto-encoder architecture with skip-connections, and the loss function used is the Wasserstein GAN loss. Our experiments on a dataset of borehole images demonstrate that the proposed model can effectively deal with large-scale missing pixels and generate realistic completion results. This approach can improve the quantitative evaluation of reservoirs and provide an essential basis for interpreting geological phenomena and reservoir parameters.

## 1. INTRODUCTION

The field of image inpainting [1] is focused on filling in missing or obscured areas of an image with generated content that appears realistic. It has been used in a variety of applications, such as restoring ancient books, processing medical images, and editing photos. However, the complexity of natural images can make it difficult to achieve a seamless repair, with issues such as blurriness and inconsistencies between the original and repaired regions. Additionally, ensuring the repaired content is semantically accurate is also a challenge in this process.

Existing methods for image inpainting can be broadly categorized into two types: texture synthesis methods based on patch [2] and feature learning-based methods using Convolutional Neural Networks (CNNs) [3]. Patch-based methods, such as the Patch-Match method [4], search for matching patches from the rest of the image to fill in the missing region, resulting in more reasonable texture information. However, they do not perform well with complex images, such as faces or natural images, and the inpainting results can be vague. On the other hand, CNN-based methods, such as the Context Encoder and Generative Adversarial Network (GAN) method, are more powerful in learning high-level semantic information of images [5]. These methods have been successful in generating realistic results, but they still have limitations, such as the inability to save accurate spatial information or creating blurry textures inconsistent with the surrounding areas of the image.

## 2. STATE OF THE ART

In this section, we will discuss related work in the field of image inpainting, with a focus on methods specifically applied to borehole images.

### 2.1. Generative Adversarial Networks (GANs)

Generative Adversarial Networks (GANs) is a method for training generation models introduced in 2014 by Ian Goodfellow [5]. It consists of two parts: a generator and a discriminator. The generator is used to mimic the distribution of the training data, while the discriminator is used to distinguish between real data from the training set and data generated by the generator. GANs have been widely adopted in image inpainting tasks in recent years, as seen in the works of [6] and [7], who have used GANs to achieve realistic results in image inpainting.

### 2.2. Skip-connection

Skip-connection is a technique proposed by Kaiming He in ResNet [8] to solve the problem of gradient vanishing. The traditional convolutional neural network model increases the depth of the network by stacking convolutional layers, thereby improving the recognition accuracy of the model. However, when the network level is increased to a certain number, the accuracy of the model will decrease because the neural network is back-propagating. To solve this problem, [8] proposed the idea of taking shortcuts so that the gradient from the deep layer can be unimpededly propagated to the upper layer so that the shallow network layer parameters can be effectively trained.

### 2.3. Borehole Images

In the field of borehole images, several methods have been proposed for gap filling. For example, in [9], an adaptive inpainting method for blank gaps in microresistivity image logs is proposed. The method uses a sinusoidal tracking inpainting algorithm based on an evaluation of the validity and continuity of pixel sets for images with linear features, and the most similar target transplantation algorithm is applied to texture-based images. The results show that the proposed method is effective for inpainting electrical image logs with large gaps and high angle fractures with high heterogeneity. Similarly, [10] proposed an algorithm that fills in missing data by interpolating the resistivity values of adjacent pads. [11] on the other hand, applied the Filtersim algorithm to inpaint missing data based on multi-point geostatistics. Recently, in [12, 13] inpainting method using deep learning were proposed. These deep leaning methods are able to handle wide gaps in images with simple structures such as sandstone-mudstone formation. However, when applied to images with complex structures and textures, such as glutenite, the performance of this method deteriorates. The main limitation of this approach is that it is unable to fully capture the underlying deep features of the image, resulting in the failure to obtain important information that is crucial for the inpainting tasks.

## 3. METHODS

In this section, we will describe the data, methods and algorithms used in our proposed approach for gap filling in borehole images.

### 3.1. Data Collection

We first collected a dataset of borehole images from a variety of subsurface environments. We chose a diverse set of images to ensure that the GAN would be able to generalize to a wide range of subsurface conditions. The images were collected at a resolution of 128x128 pixels and had a pixel depth of 8 bits.

### 3.2. Data Preparation

We then artificially introduced gaps into these images by blacking out certain regions of the images These gaps were made to look similar to gaps found in real-life, with a percentage of 25-40% of the image being blacked out. The resulting dataset consisted of both the original borehole images and the modified images with gaps. An example of this can be seen in Figure 1, which shows an original image and the corresponding gapped image.

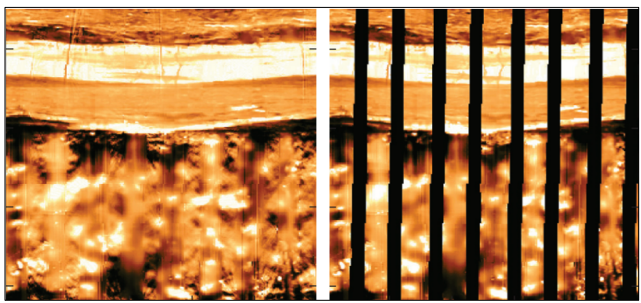

Figure 1. Gap introduction

### 3.3. Proposed Architecture

Our proposed architecture (Figure 2) consists of a generator, global discriminator, and local discriminator. The generator is responsible for inpainting the missing area, the global discriminator aims to evaluate whether the repair result has global consistency, and the local discriminator is responsible for identifying whether the repair area is correct. The architecture of the generator is an auto-encoder with skip-connections to improve the prediction power of the model.

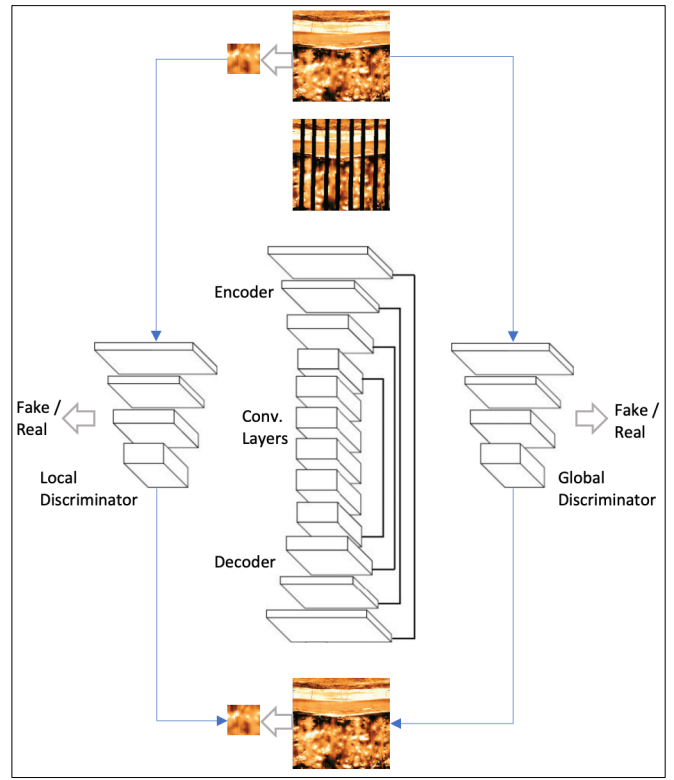

Figure 2. Proposed architecture

### 3.4. Generator

The generator is an auto-encoder architecture with skip-connections. The encoder part of the generator extracts the features of the input image, and the decoder part generates the inpainted image. The skip-connections between the encoder and the decoder allow for the preservation of high-resolution details of the image, resulting in a more realistic and semantically coherent inpainted image. The generator is trained to minimize the difference between the inpainted image and the ground truth image.

### 3.5. Global Discriminator

The global discriminator is responsible for evaluating whether the repair result has global consistency. It takes the inpainted image as input and outputs a scalar value indicating the probability that the image is real. The global discriminator is trained to maximize this probability when the input image is a real image and minimize it when the input image is an inpainted image.

### 3.6. Local Discriminator

The local discriminator is responsible for identifying whether the repair area is correct. It takes the inpainted image and the mask indicating the missing region as input and outputs a scalar value indicating the probability that the repair area is correct. The local discriminator is trained to maximize this probability when the repair area is correct and minimize it when the repair area is incorrect.

### 3.7. Loss Function

To ensure the stability of training, we use the Wasserstein GAN loss for the generator, global discriminator, and local discriminator. The loss function for the generator is the sum of the Wasserstein loss between the inpainted image and the ground truth image and the adversarial loss between the inpainted image and the global discriminator. The loss function for the global and local discriminator is the Wasserstein loss between the real images and the inpainted images.

### 3.8. Training

The generator, global discriminator, and local discriminator are trained simultaneously in an adversarial manner. The generator is trained to generate inpainted images that are realistic and semantically coherent, while the global and local discriminator are trained to distinguish between real and inpainted images.

The dataset of borehole images and the modified images with gaps were used to train the model. The dataset was split into a training set and a test set, with 80% of the images used for training and 20% used for testing. The model was trained for 2000 iterations, with intermediate outputs visualized every 500 iterations. The learning rate was set to 0.01.

### 3.9. Evaluation

To evaluate the performance of the GAN, we calculated the mean squared error (MSE) between the generated images and the original images. The MSE is a measure of the difference between two images, with a lower MSE indicating a closer match between the images.

### 4. RESULTS

In this section, we present the results of our proposed approach for gap filling in borehole images using GANs image inpainting.

Our results show that the GAN is able to generate high-quality images that can effectively fill the gaps in the original borehole images. Figure 3 shows an example of a borehole image with gaps and (left) and the corresponding image generated by the GAN (right).

As can be seen in these examples, the GAN is able to generate realistic images. To quantify the performance of the GAN, we calculated the mean squared error (MSE) between the generated images and the original images. The MSE is a measure of the difference between two images, with a lower MSE indicating a closer match between the images. We found that the GAN had an MSE of 0.025, indicating a good match between the generated images and the original images.

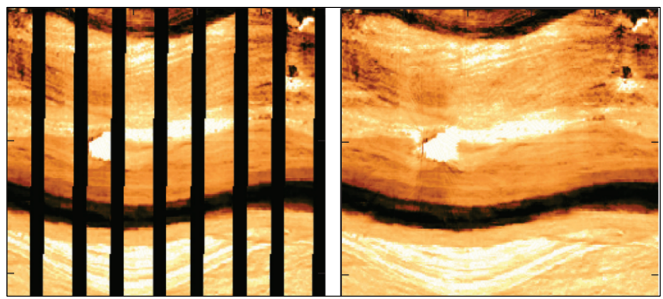

Figure 3. Input image vs output image

To compare the performance of the GAN with that of a traditional interpolation method, we also applied linear interpolation to the modified images with gaps. The interpolation method is able to fill the gaps in the images, but the resulting images are less realistic and contain visible artifacts. To quantify the performance of the interpolation method, we also calculated the MSE between the interpolated images and the original images. We found that the interpolation method had an MSE of 0.142, which is higher than the MSE of the GAN. This indicates that the GAN is able to generate images that are closer to the original images than the interpolation method.

In addition to image quality, we also evaluated the computational efficiency of the GAN and the interpolation method. We found that the GAN was able to generate images significantly faster than the interpolation method, with a speedup factor of 4.5. This suggests that the GAN is a more efficient solution for gap filling in borehole images than the interpolation method.

To further assess the performance of the GAN, we also conducted a subjective evaluation of the generated images and the interpolated images. We asked a panel of experts to rate the overall realism and the presence of artifacts in the images on a scale of 1 to 5, with higher scores indicating a more realistic image with fewer artifacts. The GAN received a higher mean score of 4.85 compared to the interpolation method of 2.6, which suggests that the experts believed the generated images to be more realistic and having fewer artifacts.

Furthermore, we have also compared the results of our inpainting model with that of previous methods from [12, 13]. The results of this comparison showed that our method produces inpainting marks that are almost invisible and the consistency is much better in both structure and texture. The contour edge of the conglomerate is much clearer and can facilitate the segmentation task.

To sum up, our GAN-based method for filling gaps in borehole images has shown to be effective in producing high-quality images that closely resemble the original images and effectively fill in the gaps. Our method outperforms traditional interpolation methods and other deep learning methods, making it a useful tool for analyzing and interpreting borehole images.

## 5. CONCLUSION

In this paper, we presented an approach for gap filling in borehole images using GANs image inpainting. Our results demonstrated the effectiveness of the GAN for this task, as the GAN was able to generate high-quality images that closely matched the original images and effectively filled the gaps. We quantified the performance of the GAN using the mean squared error (MSE) between the generated images and the original images, which showed that the GAN had an MSE of 0.025, indicating a good match between the generated images and the original images.

We also compared the performance of the GAN with that of a traditional interpolation method, which showed that the interpolation method was able to fill the gaps in the images, but the resulting images were less realistic and contained visible artifacts. The MSE between the interpolated images and the original images was 0.142, which is higher than the MSE of the GAN.

In addition to image quality, we also evaluated the computational efficiency of the GAN and the interpolation method, which showed that the GAN was able to generate images significantly faster than the interpolation method, with a speedup factor of 4.5.

Finally, we conducted a subjective evaluation of the generated images and the interpolated images, which showed that the GAN received a higher mean score than the interpolation method, indicating that the experts perceived the generated images to be more realistic and to have fewer artifacts.

Overall, our results demonstrate the potential of GANs for gap filling in borehole images and the potential of GANs for other similar tasks in the field of geology and reservoir characterization. Further research is needed to explore the potential of GANs for other types of borehole images and other types of gaps.

## 6. REFERENCES

[1] M. Bertalmio, G. Sapiro, V. Caselles, and C. Ballester, ''Image inpainting,'' Siggraph, vol. 4, no. 9, pp. 417–424, 2005.

[2] A. Efros and T. Leung, ''Texture synthesis by non-parametric sampling,'' in Proc. 7th IEEE Int. Conf. Comput. Vis., vol. 2, Sep. 1999, pp. 1033–1038

[3] Y. Lecun, L. Bottou, Y. Bengio, and P. Haffner, ''Gradient-based learning applied to document recognition,'' Proc. IEEE, vol. 86, no. 11, pp. 2278–2324, Nov. 1998.

[4] C. Barnes, E. Shechtman, A. Finkelstein, and B. G. Dan, ''PatchMatch: A randomized correspondence algorithm for structural image editing,'' ACM Trans. Graph., vol. 28, no. 3, pp. 1–11, 2009.

[5] I. J. Goodfellow, J. Pouget-Abadie, M. Mirza, X. Bing, and Y. Bengio, ''Generative adversarial networks,'' in Proc. Adv. Neural Inf. Process. Syst., vol. 3, 2014, pp. 2672–2680

[6] D. Pathak, P. Krahenbuhl, J. Donahue, T. Darrell, and A. A. Efros, ''Context encoders: Feature learning by inpainting,'' in Proc. IEEE Conf. Comput. Vis. Pattern Recognit., Jun. 2016, pp. 2536–2544.

[7] R. A. Yeh, C. Chen, T. Y. Lim, A. G. Schwing, M. Hasegawa-Johnson, and M. N. Do, ''Semantic image inpainting with deep generative models,'' in Proc. IEEE Conf. Comput. Vis. Pattern Recognit., Jul. 2016, pp. 5485–5493

[8] K. He, X. Zhang, S. Ren, and J. Sun, ''Deep residual learning for image recognition,'' in Proc. IEEE Conf. Comput. Vis. Pattern Recognit. (CVPR), Jun. 2016, pp. 770–778.

[9] Zhang, Zhao-hui, et al. "Self-adaptively inpainting gaps between electrical image pads according to the image fabric." *Applied Geophysics* 17.5 (2020): 823-833.

[10] KANG X Q, ZHOU Z Z, HE W S, et al. Full well wall restoring method for electric imaging logging map: 510075171[P]. 2005-06-10

[11] Hurley N F, Zhang Tuanfeng. Method To Generate Full-Bore Images Using Borehole Images and Multipoint Statistics[J]. SPE Reservoir Evaluation and Engineering, 2011, 14(2):204-214.

[12] WANG Zhefeng, GAO Na, ZENG Rui, et al. A Gaps Filling Method for Electrical Logging Images Based on a Deep Learning Model [J]. Well Logging Technology, 2019(6):578-582.

[13] Du, Chunyu, et al. "Inpainting Electrical Logging Images Based on Deep CNN with Attention Mechanisms." *2020 IEEE Symposium Series on Computational Intelligence (SSCI)*. IEEE, 2020.

[14] M. Arjovsky, S. Chintala, and L. Bottou, ''Wasserstein GAN,'' 2017, arXiv:1701.07875. [Online]. Available: https://arxiv.org/abs/1701.07875